\pdfoutput=1

\documentclass[11pt]{article}

\usepackage[preprint]{acl}

\usepackage{times}
\usepackage{latexsym}
\usepackage{graphicx}
\usepackage{booktabs}
\usepackage{soul,xcolor}
\sethlcolor{blue!30}
\newcommand{\hanqing}[1]{#1}
\newcommand{\han}[1]{\hl{#1}}
\renewcommand{\han}[1]{#1}
\newcommand{\hanq}[1]{\hl{#1}}
\renewcommand{\hanq}[1]{#1}
\usepackage[T1]{fontenc}

\usepackage[utf8]{inputenc}

\usepackage{microtype}

\usepackage{inconsolata}

\usepackage{graphicx}

\title{Zero-shot Graph Reasoning via Retrieval Augmented Framework with LLMs}

\author{
 \textbf{Hanqing Li\textsuperscript{1}},
 \textbf{Kiran Sheena Jyothi\textsuperscript{2}},
\\
 \textbf{Henry Liang\textsuperscript{3}},
 \textbf{Sharika Mahadevan\textsuperscript{4}},
 \textbf{Diego Klabjan\textsuperscript{1}}
\\
\\
 \textsuperscript{1}Northwestern University,
 \textsuperscript{2}EXL Service,
 \textsuperscript{3}Vail Systems,
 \textsuperscript{4}Netflix
\\
 \small{
   \textbf{Correspondence:} \href{mailto:hanqingli2025@u.northwestern.edu}{hanqingli2025@u.northwestern.edu}
 }
}

\begin{document}
\maketitle
\begin{abstract}
  We propose a new, training-free method, Graph Reasoning via Retrieval Augmented Framework (GRRAF), that harnesses retrieval-augmented generation (RAG) alongside the code-generation capabilities of large language models (LLMs) to address a wide range of graph reasoning tasks. In GRRAF, the target graph is stored in a graph database, and the LLM is prompted to generate executable code queries that retrieve the necessary information. This approach circumvents the limitations of existing methods that require extensive finetuning or depend on predefined algorithms, and it incorporates an error feedback loop with a time-out mechanism to ensure both correctness and efficiency. Experimental evaluations on the GraphInstruct dataset reveal that GRRAF achieves 100\% accuracy on most graph reasoning tasks, including cycle detection, bipartite graph checks, shortest path computation, and maximum flow, while maintaining consistent token costs regardless of graph sizes. Imperfect but still very high performance is observed on subgraph matching. Notably, GRRAF scales effectively to large graphs with up to 10,000 nodes.
\end{abstract}

\section{Introduction}

\begin{figure}[t]
    \centering
    \includegraphics[width=\columnwidth]{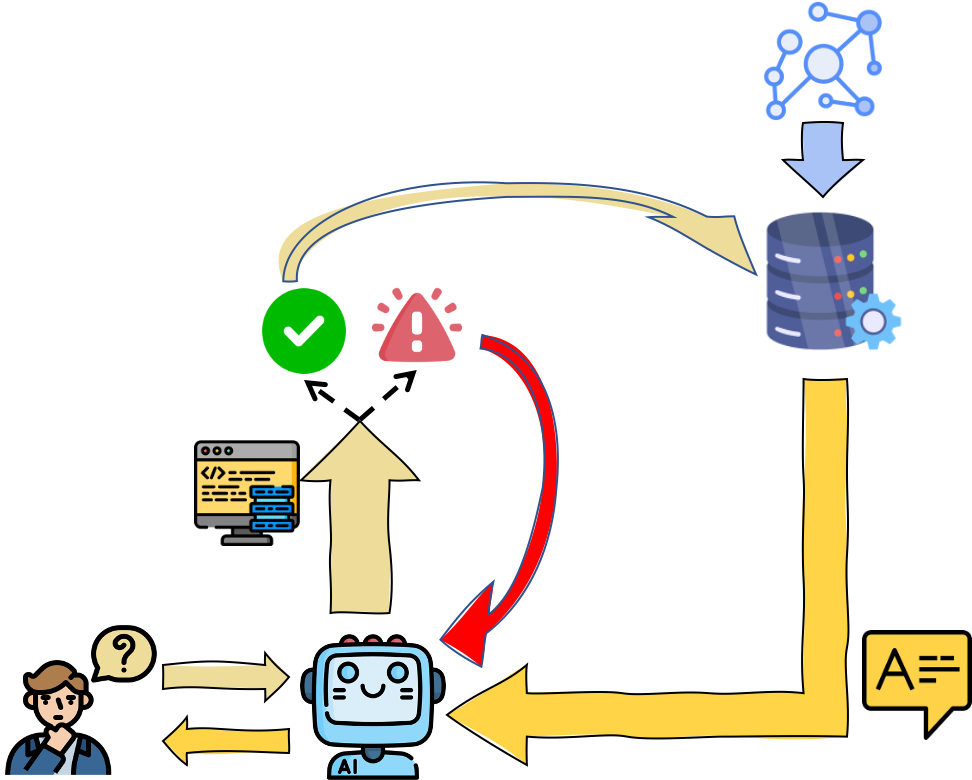}
    \caption{A schematic representation of the GRRAF concept. When a user asks a graph reasoning question, the LLM generates code to query the target graph stored in a graph database, retrieves the answer, and presents it as the response. An error feedback loop is integrated into GRRAF to prompt the LLM to refine the code whenever execution or time-out errors occur.}
    \label{fig:simple_idea}
\end{figure}
Graph reasoning plays a pivotal role in modeling and understanding complex systems across numerous domains \cite{wu2020comprehensive}. Graphs naturally represent entities and their interrelations in areas such as social networks, transportation systems, biological networks, and communication infrastructures. Graph reasoning tasks like determining connectivity, detecting cycles, and finding the shortest path are not only central to theoretical computer science but also have practical implications in network optimization, anomaly detection, decision support systems, etc \cite{scarselli2008graph}. However, addressing these tasks requires a deep understanding of graph topology combined with precise computational procedures, underscoring the critical challenge of developing efficient graph reasoning methods in contemporary machine learning research \cite{zhao2024graphtext}.

Large language models (LLMs) have demonstrated an impressive capacity for multi-step reasoning, which enables them to interpret complex graph-related questions expressed in natural language and generate human-readable responses \cite{guo2023gpt4graph}. Several recent studies have leveraged LLMs to tackle graph reasoning problems by converting graph structures into textual representations or latent embeddings through graph neural networks (GNNs), thereby exploiting the powerful natural language reasoning capabilities of LLMs \cite{perozzi2024let,guo2023gpt4graph,zhang2023graph,wang2024can,fatemitalk,skianis2024graph,lin2024langgfm}. However, even when advanced prompting techniques are employed, these methods tend to perform poorly on fundamental graph reasoning tasks, such as evaluating connectivity or identifying the shortest path, with average accuracies ranging from 20\% to 60\%. Alternative approaches that achieve higher accuracy typically either require extensive finetuning—which results in poor performance on out-of-domain questions \cite{chen2024graphwiz,zhang2023graph}—or rely on predefined algorithms as input, thereby limiting their ability to address unseen tasks \cite{hu2024scalable}.

To address these limitations, we introduce a training-free and zero-shot method, \textit{\textbf{\hanq{the} Graph Reasoning via Retrieval Augmented Framework}} (GRRAF), that leverages retrieval-augmented generation (RAG) \cite{lewis2020rag} alongside the code-writing capabilities of large language models. In GRRAF, the target graph is stored in a graph database, and the LLM is prompted to generate appropriate queries, written as code, that extract the desired answer by retrieving relevant information from the database. This strategy harnesses the LLM’s robust reasoning ability and its proficiency in generating executable code, thereby achieving high accuracy on a range of graph reasoning tasks without requiring additional finetuning or predefined algorithms. In addition, we incorporate an error feedback loop combined with a time-out mechanism to ensure that the LLM produces correct queries in a time-efficient manner. Furthermore, since accurate code reliably yields the correct answer regardless of the graph's size, GRRAF can easily scale for polynomial problems to accommodate larger graphs without a drop in accuracy. In GRRAF, we use Neo4j, an interactive graph database, and NetworkX, a Python library for graphs. GRRAF accepts the target graph as \han{either} plain text \han{or} data already stored in Neo4j, \han{specified in the prompt by the graph file name. In the former case, the prompt must specify if Neo4j or NetworkX is to be used. The LLM then must either create code to insert the graph specified in the prompt to Neo4j or to a NetworkX graph object.} 

GRRAF offers a fully automated, end-to-end framework for handling graph-reasoning problems written entirely in text. By leveraging the world knowledge encoded in LLMs, it generates correct code and returns accurate answers automatically for a wide range of graph-reasoning tasks expressed as natural-language questions. In addition, GRRAF establishes a foundation for future work on real-world structured relational-inference problems—ranging from knowledge-graph completion to molecular analysis—that are naturally represented as graph-structured data. \han{An LLM user could potentially accomplish the same by directly prompting the LLM to create Python or Neo4j queries for the task on hand. Our approach offers the benefits of graph reading and loading, the execution of the code with the error-feedback loop, and the fallback approach.}

Experimental results demonstrate that GRRAF achieves 100\% accuracy on \hanqing{many} graph reasoning tasks, outperforming state-of-the-art benchmarks. Moreover, GRRAF is applicable to large graphs containing up to 10,000 nodes, maintaining 100\% accuracy with no increase in token cost. \hanqing{Although GRRAF only achieves 86.5\% accuracy on subgraph matching, it still outperforms other state-of-the-art methods.} Our contributions are listed below.
\begin{itemize}
    \item \textbf{Novel Graph Reasoning Approach:} This work introduces a \hanqing{new} method that leverages RAG to address graph reasoning tasks, such as connectivity analyses, cycle detection, and shortest path computations. It represents the first application of RAG in the domain of graph reasoning.
    \item \textbf{Error Feedback Loop Innovation:} The paper introduces the integration of a time out mechanism within an error feedback loop, along with the dynamic refreshing of a prompt to guide the LLM to produce more efficient code. This mechanism enhances robustness and efficiency of the generated query by preventing an infinite loop.
    \item \textbf{Scalable State-of-the-Art Performance:} The proposed method achieves state-of-the-art accuracy and demonstrates exceptional scalability, being the first to handle large graphs effectively without significant degradation in accuracy or substantial cost increases.
\end{itemize}

All implementations and datasets are available in \url{https://github.com/hanklee97121/GRRAF/tree/main}.

\section{Related Works}
\label{sec:related}
\begin{figure*}[t]
    \centering
    \includegraphics[width=\linewidth]{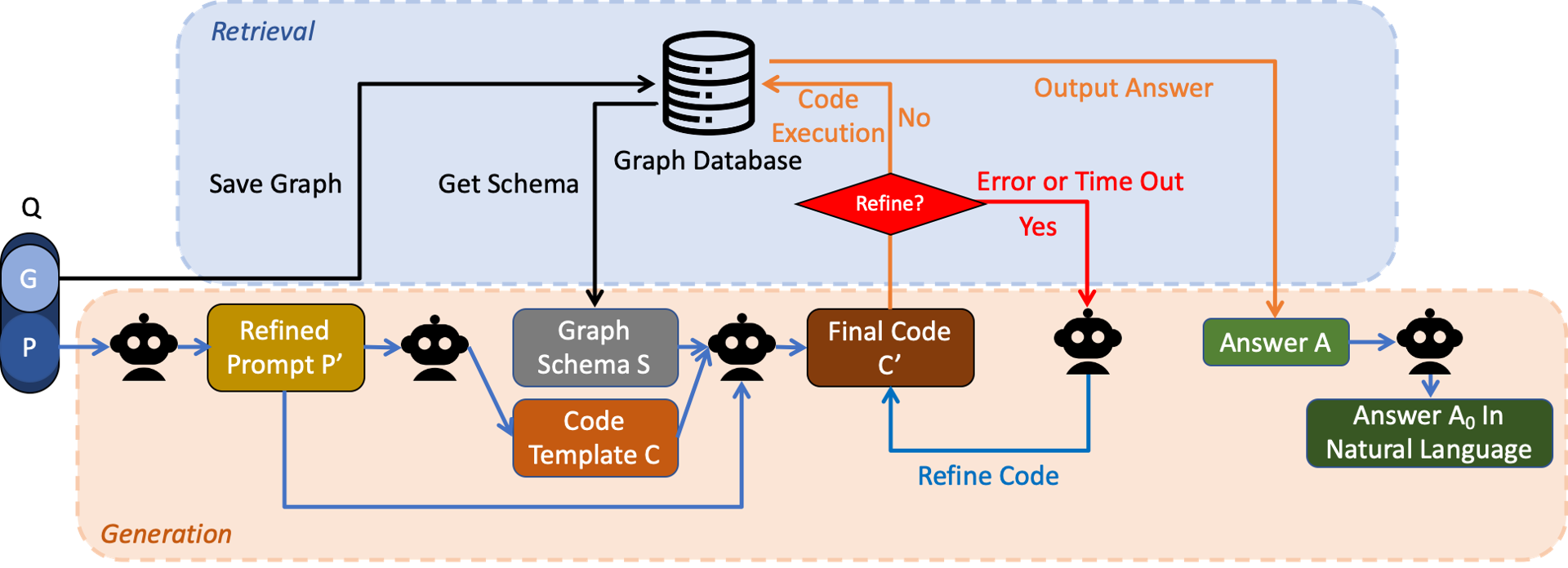}
    \caption{\hanqing{GRRAF workflow. The retrieval component represents the interaction with the graph database through code, while the generation component involves prompting an LLM to produce the output.}}
    \label{fig:workflow}
\end{figure*}
\subsection{Graph RAG}
There exist numerous prior works that employ graph data within RAG frameworks to enhance the capabilities of LLMs, a paradigm often referred to as GraphRAG \cite{peng2024graph}. These approaches retrieve graph elements containing relational knowledge relevant to a given query from a pre-constructed graph database \cite{edge2024local}. Several studies have contributed to the development of open-source knowledge graph datasets for GraphRAG \cite{auer2007dbpedia,suchanek2007yago,vrandevcic2014wikidata,sap2019atomic,liu2004conceptnet,bollacker2008freebase}. Building on these datasets, many methods opt to convert graphs \hanqing{to} other easily retrievable forms, such as text \cite{li2023graph,huang2023mvp,yu2023decaf,edge2024local,dehghan-etal-2024-ewek} or vectors \cite{he2024gretriever,sarmah2024hybridrag}, to improve the efficiency of query operations on graph databases. To further enhance the quality of retrieved data, several approaches optimize the retrieval process within GraphRAG by refining the retriever component \cite{anonymous2024graphbased,zhang-etal-2022-subgraph,kim2023kg,wold-etal-2023-text,jiang-etal-2023-structgpt,mavromatis2024gnn}, optimizing the retrieval paradigm \cite{wang2024knowledge,sun-etal-2024-oda,lin-etal-2019-kagnet}, and editing a user query or the retrieved information \cite{jin2024graphcot,luo2024reasoning,ma2025thinkongraph,sunthink,taunk2023grapeqa,yasunaga2021qa}. Furthermore, many methods enhance the answer generation process of GraphRAG to ensure that the LLM fully utilizes the retrieved graph data to generate the correct answer \cite{dong2023hierarchy,mavromatis2022rearev,jiang2024reasoning,sun2024thinkongraph,zhang2022greaselm,zhu2024efficient,wen-etal-2024-mindmap,shu-etal-2022-tiara,baek-etal-2023-knowledge-augmented-language}. However, these methods focus exclusively on knowledge graphs and cannot be directly applied to solve graph reasoning questions. In contrast, GRRAF is the first method to employ RAG for addressing graph reasoning questions on pure graphs.

\subsection{Graph Reasoning}
\label{sec:graph_reasoning}
Recent work has explored the use of LLMs to address graph reasoning problems. Several methods rely solely on prompt engineering techniques to enhance LLM reasoning capabilities on graphs \cite{liu2023evaluating,guo2023gpt4graph,wang2024can,zhang2024dyg,fatemitalk,wu2024grapheval2000,tang2025grapharena,skianis2024graph,lin2024langgfm}. Building on them, \citet{perozzi2024let} integrate a trained graph neural network \cite{scarselli2008graph} with an LLM to improve its performance on graph reasoning tasks by encoding each graph into a token provided as input to the LLM. Meanwhile, \citet{zhang2023graph} and \citet{chen2024graphwiz} finetune an LLM with instructions tailored to graph reasoning tasks to boost performance. In another approach, \citet{hu2024scalable} propose a multi-agent solution for graph reasoning problems by assigning an LLM agent to each node and enabling communication among agents based on a predefined algorithm. In contrast, GRRAF employs RAG to address graph reasoning problems \hanqing{without extensive prompt engineering.} This approach is training-free \hanqing{and thus unsupervised} and does not depend on any predefined algorithm. Furthermore, unlike previous methods, the LLM in GRRAF does not receive the entire graph as input; consequently, the token usage remains independent of graph size, thereby enabling efficient scalability to very large graphs.
\begin{figure*}[t]
    \centering
    \includegraphics[width=\linewidth]{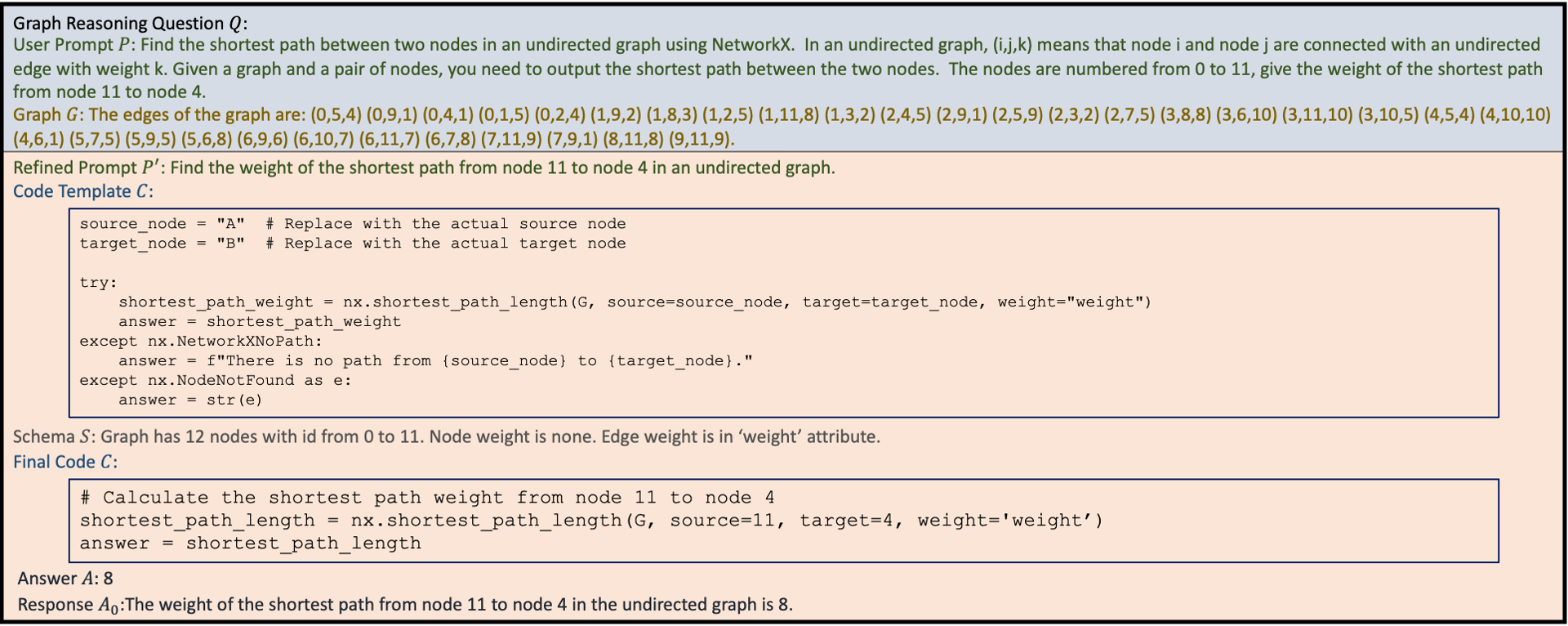}
    \caption{An illustrative example demonstrating the application of GRRAF to solve a shortest path question by using NetworkX. Graph $G$ in text is stored as an NetworkX object by code.}
    \label{fig:example}
\end{figure*}

\section{Method}
\label{sec:method}

\begin{table}[t]
  \centering
  \begin{tabular}{|l|l|p{1.1cm}|}
    \hline
    \textbf{Task}           & \hanqing{\textbf{\han{Node} Size}} & \hanqing{\textbf{\# of \han{Test} Graphs}}\\
    \hline
    Cycle Detection&[2, 100]&400\\
    \hline
    Connectivity&[2, 100]&400\\
    \hline
    Bipartite Graph Check&[2, 100]&400\\
    \hline
    Topological Sort&[2, 50]&400\\
    \hline
    Shortest Path&[2, 100]&400\\
    \hline
    Maximum Triangle Sum&[2, 25]&400\\
    \hline
    Maximum Flow&[2, 50]&400\\
    \hline
    Subgraph Matching&[2, 30]&400\\
    \hline
    Indegree Calculation&[2, 50]&400\\
    \hline
    Outdegree Calculation&[2, 50]&400\\
    \hline
  \end{tabular}
  \caption{
    The detailed information of GraphInstruct dataset and two additional tasks (indegree calculation and outdegree calculation). \hanqing{The subgraph matching task is to verify if there exists a subgraph in $G$ that is isomorphic to a given graph $G'$.}
  }
  \label{tab:dataset}
\end{table}
In this section, we explain how GRRAF integrates RAG to address graph reasoning questions and retrieve accurate answers. The entire workflow of GRRAF is demonstrated in Figure \ref{fig:workflow}. A graph reasoning question, denoted as $Q$, consists of two components: a graph $G$ and a user prompt $P$. The graph $G$ represents the target graph associated with $Q$ and is stored \han{either in Neo4j or as a NetworkX graph object (code written by an LLM and executed by an agent).} The prompt $P$ contains a graph-specific question regarding $G$ (e.g., ``Does node 2 connect to node 5?'' or ``What is the shortest path from node 5 to node 8?''). To enhance code generation by the language model, we initially input $P$ into the LLM, requesting it to refine the prompt, clarify the format, and eliminate redundant information. The resulting refined prompt is denoted as $P'$. Then, the LLM is prompted to generate a generic code template $C$ that addresses $P'$ without incorporating graph-specific details. For example, if $P'$ states ``Find the shortest path from node 3 to node 5,'' the template $C$ encapsulates a generic shortest path algorithm that does not include the specific node identifiers. Additionally, we extract the schema $S$ (comprising of node properties and edge properties) from the graph database using a hard-coded procedure. This schema ensures that the LLM-generated code utilizes correct variable names.

Subsequently, we provide $P'$, $C$, and $S$ to the LLM and instruct it to generate the final code $C'$ that produces \hanqing{an} answer $A$ corresponding to $P'$. An error feedback loop is incorporated into this process. If an error arises during the execution of $C'$, the error message, along with $C'$, is supplied back to the LLM, prompting it to produce a revised version of the code. To promote the generation of time-efficient code, given that multiple algorithms with varying time complexities may be applicable, we integrate a time-out mechanism within the error feedback loop. Specifically, a time limit \hanqing{$t$} is imposed on the execution of $C'$. If the execution time exceeds \hanqing{$t$}, the process is halted, and the LLM is asked to modify $C'$ so that it runs faster. If the feedback loop iterates more than $n$ times, the system reverts to using the original question $Q$ as a prompt to directly obtain the answer $A$ from the LLM. This forced exit is designed to prevent perpetual iterations when addressing computationally intractable NP-hard problems (e.g., substructure matching on large graphs), where no modification of $C'$ can reduce the execution time below the threshold \hanqing{$t$}.

In the final step, the answer $A$ is provided to the LLM to generate a reader-friendly natural language response $A_0$ that addresses the graph reasoning question $Q$. An example of solving a graph reasoning question with GRRAF is demonstrated in Figure \ref{fig:example}.

\section{Computational Assessment}
\label{sec:exp_ass}
\subsection{Dataset and Benchmark}
\label{sec:data}
\begin{figure*}[t]
    \centering
    \includegraphics[width=\linewidth]{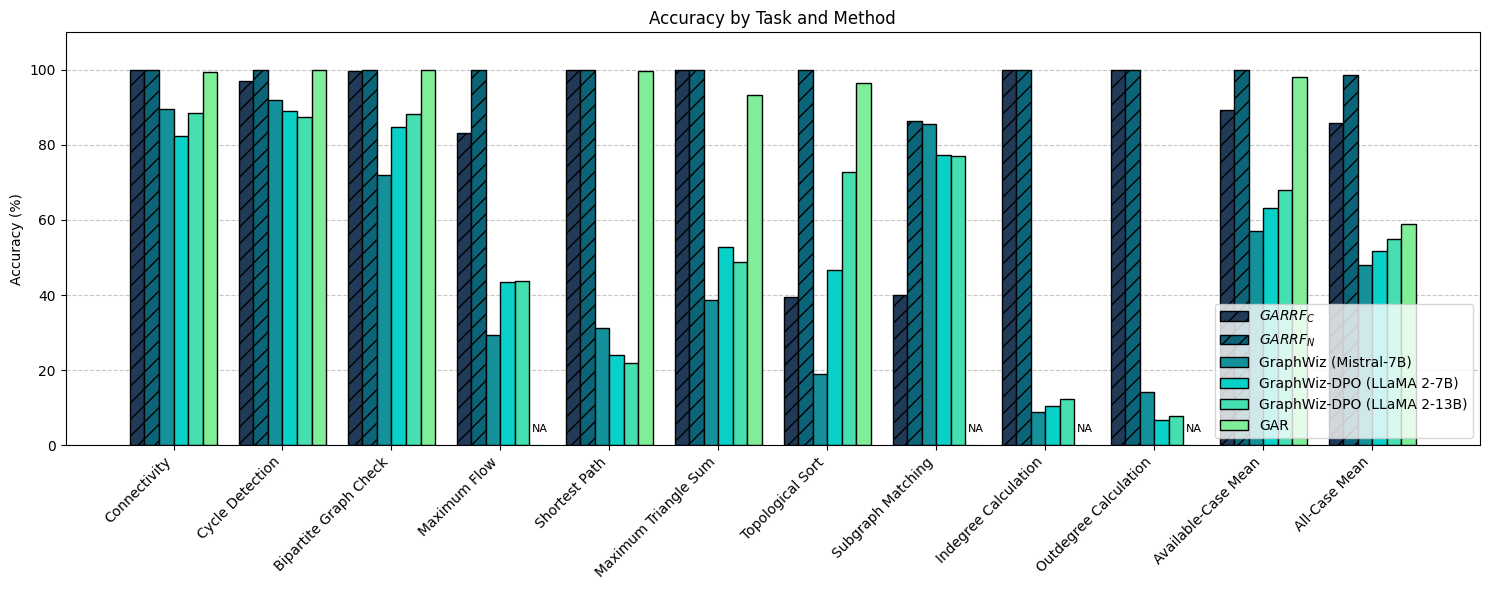}
    \caption{Performance of GRRAF and benchmark models across all ten graph reasoning tasks. Missing data are indicated as ``NA'' in the plot. \hanqing{The available-case mean refers to the average accuracy of each method calculated using only the tasks where complete data is available (excluding maximum flow, subgraph matching, indegree calculation, and outdegree calculation). The all-case mean refers to the average accuracy across all tasks, treating 'NA' as 0.}}
    \label{fig:all_result}
\end{figure*}
We conduct experiments on GraphInstruct \cite{chen2024graphwiz}, a dataset that comprises \hanqing{of} nine graph reasoning tasks with varying complexities. Due to its diversity in graph reasoning tasks and its prior use in evaluating state-of-the-art methods \cite{chen2024graphwiz,hu2024scalable}, we select this dataset for our evaluation. However, the task of finding a Hamilton path lacks publicly available ground truth labels and generating such labels through code is infeasible due to the NP-hard nature of the problem; consequently, we exclude this task from our experiments. Accordingly, we assess the performance of GRRAF on the following eight tasks: cycle detection, connectivity, bipartite graph check, topological sort, shortest path, maximum triangle sum, maximum flow, and subgraph matching. Details of these tasks are provided in Table \ref{tab:dataset}. Moreover, to achieve a more robust performance evaluation, we augment the test dataset with two additional simple tasks—indegree calculation and outdegree calculation (as shown in Table \ref{tab:dataset})—to facilitate a comprehensive evaluation of GRRAF and the state-of-the-art benchmarks. Each task has 400 question–graph pairs, each with a single correct answer. We measure a method’s performance on one task by its accuracy—that is, the proportion of questions answered correctly out of the total (400).

GRRAF, i.e., its LLM, generates code which is either correct or not. This is the reason why most \hanq{accuracies} are going to be 100\%. For tasks with less than 100\% accuracy, GRRAF yields correct code but the underlying problems are NP-hard and for some test graphs the execution times out. One can argue that the output code is correct and thus appropriate credit should be given, but on the other hand, a more efficient algorithm and code can be potentially produced. Sometimes the generated code does not handle edge cases correctly, yet other times the code or algorithms are incorrect (they solve only some test graphs by coincidence).

We compare the performance of GRRAF against two state-of-the-art benchmarks: GraphWiz \cite{chen2024graphwiz} and GAR \cite{hu2024scalable}. GraphWiz is trained on 17,158 questions and 72,785 answers, complete with reasoning paths, from the training set of GraphInstruct. Since no single version of GraphWiz consistently outperforms the others across all tasks, we include three versions in our comparisons: GraphWiz (Mistral-7B), GraphWiz-DPO (LLaMA 2-7B), and GraphWiz-DPO (LLaMA 2-13B). GAR is a training-free multi-agent framework that relies on a predefined library of distributed algorithms created by humans. As a result, it is incapable of solving unseen graph reasoning tasks that require algorithms not present in its library. Therefore, some results from GAR are missing in the subsequent comparisons because of its limitation.

\subsection{Experiments}
We conduct experiments using GRRAF with a time limit of $t = 5$ minutes and a maximum error feedback loop iteration of $n = 3$. The backbone LLM is GPT-4o. These parameter choices are justified by the sensitivity analysis in Appendix \ref{app:ablation}. For the graph querying code, we evaluate two approaches: Cypher, a query language for Neo4j, and NetworkX, a Python library for graphs, which we denote as GRRAF$_{C}$ and GRRAF$_{N}$, respectively. \han{We deal with graph plain text, and thus can be converted into either Neo4j data or NetworkX objects.}

Figure \ref{fig:all_result} demonstrates that GARRF$_N$ outperforms all benchmark methods, achieving 100\% accuracy on most graph reasoning tasks. GARRF$_C$ exhibits comparable or superior performance relative to other benchmarks on the majority of tasks, except for topological sort and subgraph matching. Although GraphWiz outperforms GARRF$_C$ in topological sort and subgraph matching, its inadequate performance on indegree calculation and outdegree calculation suggests that it struggles with even simple out-of-domain graph reasoning tasks. Furthermore, due to its inherent limitations, GAR is inapplicable to out-of-domain tasks such as maximum flow, subgraph matching, indegree calculation, and outdegree calculation. Consequently, considering both performance and generalization ability, GARRF$_C$ and GARRF$_N$ are better for addressing graph reasoning tasks than the other benchmark models. \hanqing{The example code generated by GARRF$_N$ for each graph reasoning task is presented in Appendix} \ref{app:code_example}.

\han{Subgraph matching is NP-complete, and the code produced by GARRF$_N$ has exponential time complexity. For graphs of 20 nodes, executing that code can take over a day—exceeding the time limit $t$.} Based on Section \ref{sec:method}, in such cases GARRD$_N$ falls back to using the original question $Q$ as a prompt to obtain the answer $A$ directly from the LLM, which may yield incorrect results. GRRAF$_C$ likewise falls short of 100\% accuracy on cycle detection and bipartite-graph checking, since Cypher queries execute more slowly than NetworkX. For the maximum-flow task, GRRAF$_C$ produces code that overlooks certain edge cases. And for topological sorting and subgraph matching, it generates code that only succeeds on some graphs by chance.

Across \hanqing{the} ten tasks, solving a single graph reasoning question requires GRRAF$_N$ to use an average of 767 input tokens and 124 output tokens, while GRRAF$_C$ uses 796 input tokens and 201 output tokens. In comparison, GraphWiz (Mistral-7B) consumes an average of \hanqing{1,046} input tokens and 126 output tokens per question, whereas GraphWiz-DPO (LLaMA 2-7B) requires \hanqing{1,046} input tokens and 290 output tokens on average, and GraphWiz-DPO (LLaMA 2-13B) uses \hanqing{1,046} input tokens and 301 output tokens per question. Notably, GAR demands more resources, averaging 8,095 input tokens and 5,987 output tokens for each graph reasoning question. Thus, comparing to other benchmark methods, GRRAF$_N$ and GRRAF$_C$ achieve high accuracy in graph reasoning tasks while utilizing fewer token resources.

\begin{figure}
    \centering
    \includegraphics[width=\columnwidth]{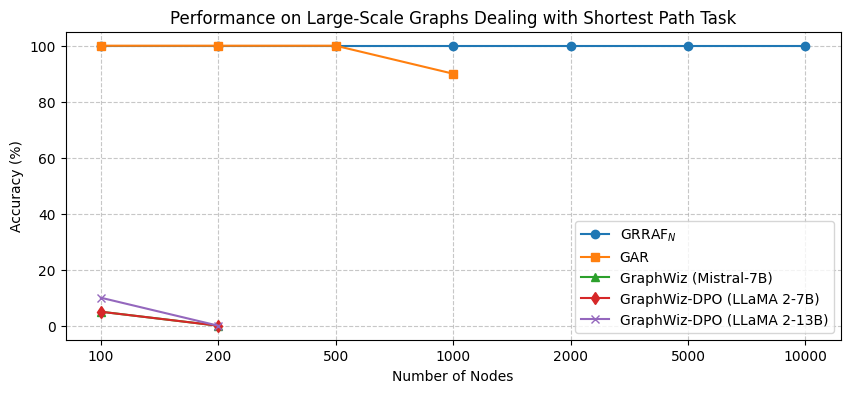}
    \caption{Accuracy of each method on the shortest path task across graphs of differenct sizes (number of nodes).}
    \label{fig:large_accuracy}
\end{figure}
Since the largest graph in GraphInstruct \cite{chen2024graphwiz} comprises \hanqing{of} only 100 nodes, which remains insufficient for real-world graph reasoning scenarios \cite{hu2024scalable}, we further evaluate the best-performing method, GRRAF$_N$, on large-scale graphs. Following the approach of \citet{hu2024scalable}, we assess GRRAF$_N$ on the shortest path task using larger graphs with 20 test samples for each graph size. Whereas their work scales graphs to 1,000 nodes, we extend this evaluation by scaling graphs to 10,000 nodes to thoroughly assess the performance of GRRAF$_N$. According to Figure \ref{fig:large_accuracy}, GRRAF$_N$ achieves 100\% accuracy across all graph sizes, demonstrating its exceptional scalability. GAR attains 100\% accuracy on graphs with 100, 200, and 500 nodes, but its accuracy decreases to 90\% on graphs with 1,000 nodes. Due to token limitations, GAR is unable to address questions on graphs with 2,000 nodes or more. In contrast, all three versions of GraphWiz perform poorly on large graphs, achieving only 5-10\% accuracy on graphs with 100 nodes and failing entirely on graphs with 200 nodes. The token limits of their base model prevent them from processing graphs larger than 200 nodes.

\begin{figure}
    \centering
    \includegraphics[width=\columnwidth]{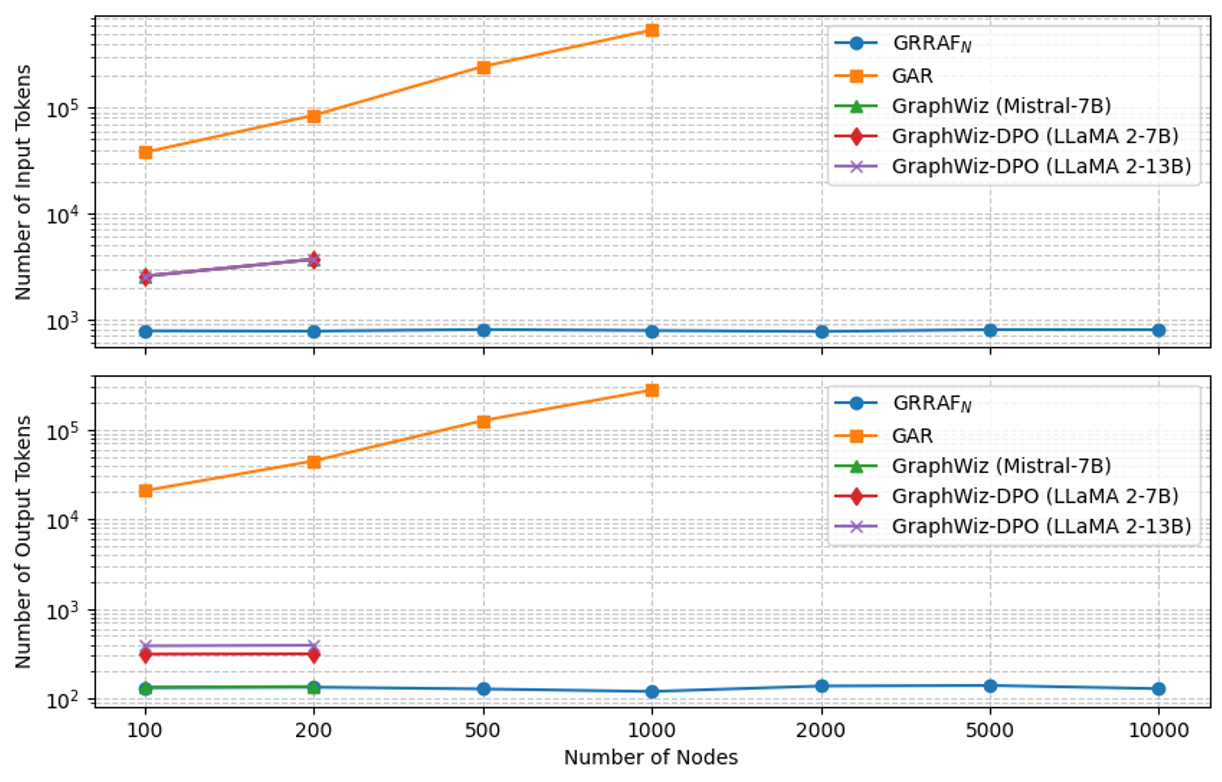}
    \caption{Average token cost for solving a graph reasoning problem across graphs of varying sizes \hanqing{on the shortest path task.}}
    \label{fig:token}
\end{figure}
We also record the variation in token cost required to solve a single graph reasoning question as the graph size increases \hanqing{on the shortest path task.} As illustrated in Figure \ref{fig:token}, the number of tokens used by GRRAF$_N$ remains constant regardless of the graph size. As detailed in Section \ref{sec:method}, GRRAF interacts with the graph solely via the graph database through code execution; thus, the graph description (nodes, edges, weights) is not directly input to the LLM, and the token cost remains unaffected by increases in graph size. In contrast, the token cost for GraphWiz increases linearly with graph size because it must pass the information of each node and edge to the LLM. The token cost for GAR is considerably higher than that for GRRAF$_N$ and grows nearly exponentially with graph size. This is due to GAR's design, where each node is assigned an LLM agent, and each agent communicates with every adjacent agent in each iteration \cite{hu2024scalable}. As the number of nodes increases, so do the number of agents, the number of adjacent agents per node (i.e., edges), and the number of iterations required to obtain an answer, all of which contribute to a significant rise in token cost. Therefore, compared to other benchmarks, GRRAF$_N$ can readily scale to very large graphs (up to 10,000 nodes) without compromising performance and increasing token cost.

\begin{table}[t]
\centering
\begin{tabular}{lcc}
\toprule
\textbf{Method} & \textbf{Execution Error} & \textbf{Time-out} \\
\midrule
GRRAF$_N$  & 2.2\%  & 5.4\% \\
GRRAF$_C$  & 4.9\%  & 9.1\% \\
\bottomrule
\end{tabular}
\caption{Percentage of graph reasoning questions over 10 tasks triggering error feedback loop due to execution errors or time-outs for each method.}
\label{tab:error_feedback}
\end{table}

To evaluate the effectiveness of the error feedback loop, we quantify the total percentage of questions that activate this loop, as reported in Table \ref{tab:error_feedback}. In general, GRRAF$_C$ triggers the error feedback loop more frequently than GRRAF$_N$. For both variants, the loop is activated due to time-outs more often than due to execution errors, underscoring the importance of time efficiency in graph reasoning tasks. Overall, the backbone LLM generates correct code queries in most instances, and the integration of an error feedback loop with a time-out mechanism further enhances code accuracy and efficiency.

\subsection{Sensitivity Analysis}
\label{app:ablation}
\begin{figure}[h]
    \centering
    \includegraphics[width=\columnwidth]{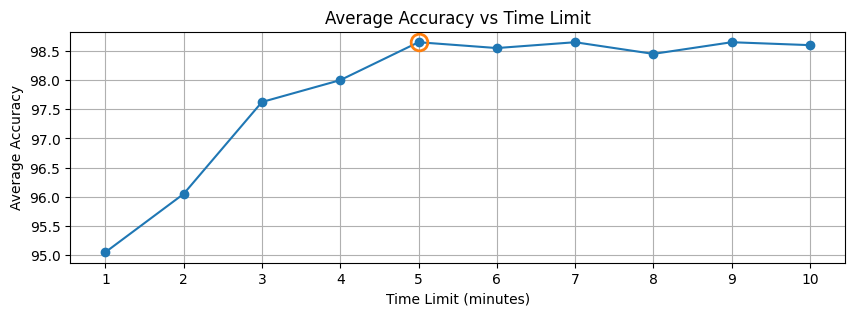}
    \caption{\hanqing{Average accuracy of GRRAF$_N$ with different time limit $t$.}}
    \label{fig:t_ablation}
\end{figure}

\begin{figure}[h]
    \centering
    \includegraphics[width=\columnwidth]{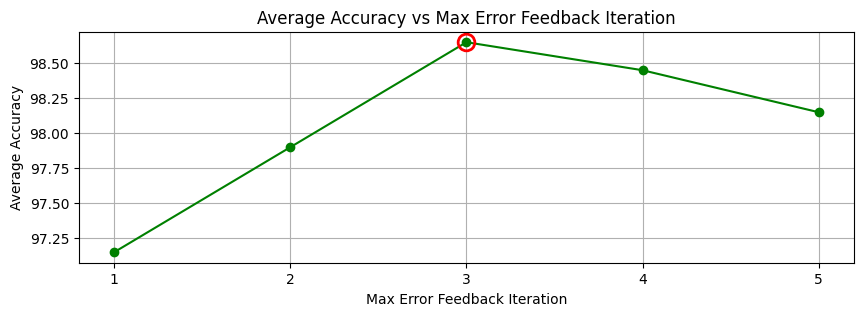}
    \caption{\hanqing{Average accuracy of GRRAF$_N$ with different maximum error feedback loop iteration $n$.}}
    \label{fig:n_ablation}
\end{figure}

\begin{figure}[h]
    \centering
    \includegraphics[width=\columnwidth]{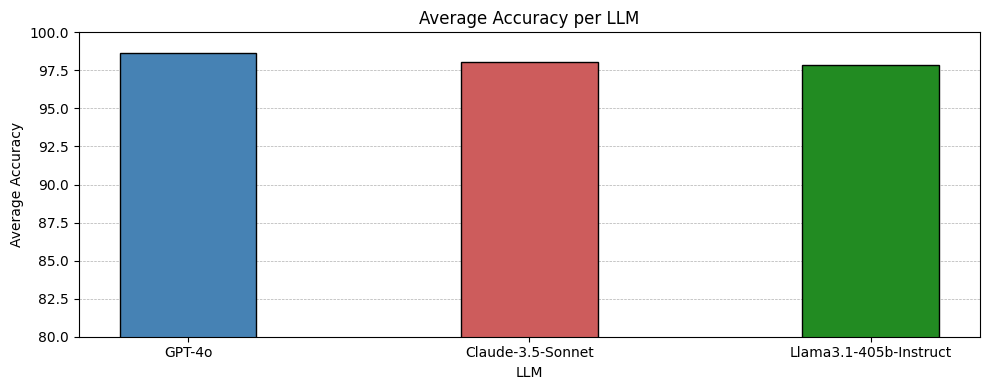}
    \caption{\hanqing{Average accuracy of GRRAF$_N$ with different backbone LLM.}}
    \label{fig:LLM_ablation}
\end{figure}
We perform sensitivity analyses on GRRAF$_N$ to assess the impact of the time limit $t$, the maximum number of error-feedback loop iterations $n$, and the choice of backbone LLM. We report the average accuracy across all ten graph reasoning tasks. As shown in Figure \ref{fig:t_ablation}, \han{the} accuracy increases with $t$ up to five minutes, after which no further gains are observed. Figure \ref{fig:n_ablation} indicates that accuracy peaks at $n=3$ and declines slightly for $n>3$. Finally, we evaluated GRRAF$_N$ using three backbone LLMs—GPT-4o, Claude-3.5-Sonnet, and Llama3.1-405b-Instruct—and found that all three yield comparable results, with GPT-4o achieving a slightly higher average accuracy than the others (Figure \ref{fig:LLM_ablation}).

\section{Conclusion}
\label{sec:conclusion}
In this work, we introduced GRRAF, a novel framework that integrates RAG with the code-writing prowess of LLMs to address graph reasoning questions. Our approach, which operates without additional training or reliance on predefined algorithms, leverages a graph database to store target graphs and employs an error feedback loop with a time-out mechanism to ensure the generation of correct and efficient code queries. Comprehensive experiments on the GraphInstruct dataset and two extra tasks (indegree and outdegree) demonstrate that GRRAF outperforms existing state-of-the-art benchmarks, achieving 100\% accuracy on a majority of graph reasoning tasks while effectively scaling to graphs containing up to 10,000 nodes without incurring extra token costs. These findings underscore the potential of combining retrieval-based techniques with LLM-driven code generation for solving complex graph reasoning problems. Future work could explore extending this framework to dynamic graph scenarios and additional reasoning tasks, further enhancing its applicability and robustness.

\section{Limitations}
Although GRRAF$_N$ attains 100\% accuracy on all polynomial‐time graph reasoning tasks, it nevertheless struggles to solve NP‐complete problems—such as subgraph matching—both accurately and efficiently. Moreover, the inferior performance of GRRAF$_C$ relative to GRRAF$_N$ indicates that our framework currently generates lower‐quality Cypher queries than the equivalent Python code. These two issues constitute the primary limitations of our method.

\bibliography{main}

\appendix

\section{Example Code}
\label{app:code_example}

\begin{figure*}
    \centering
    \includegraphics[width=\linewidth]{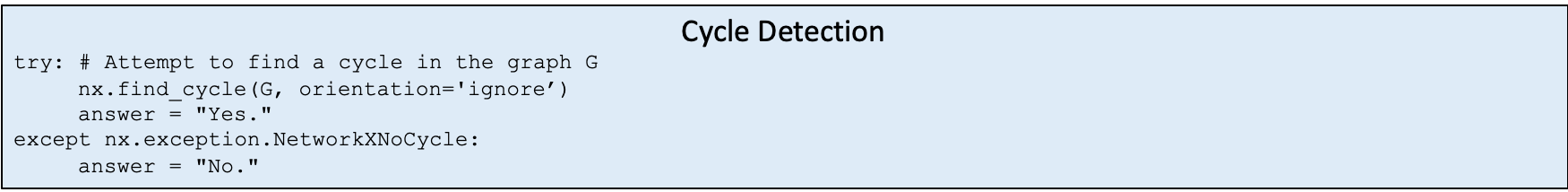}
    \caption{An example of the final code $C'$ generated for the cycle detection task.}
    \label{fig:cycle_exp}
\end{figure*}

\begin{figure*}
    \centering
    \includegraphics[width=\linewidth]{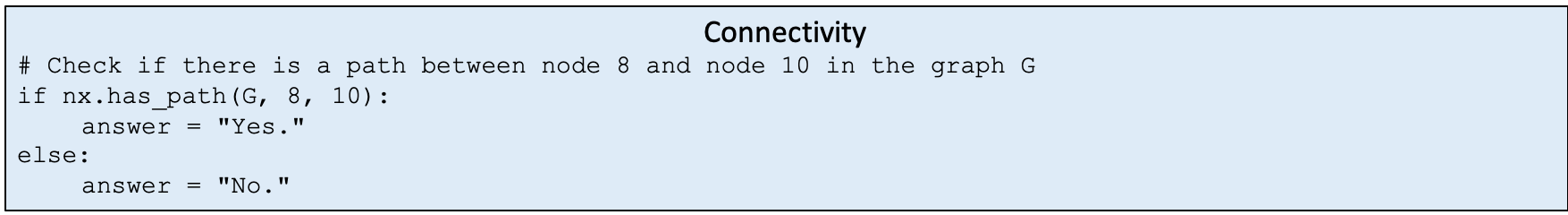}
    \caption{An example of the final code $C'$ generated for the connectivity task.}
    \label{fig:connect_exp}
\end{figure*}

\begin{figure*}
    \centering
    \includegraphics[width=\linewidth]{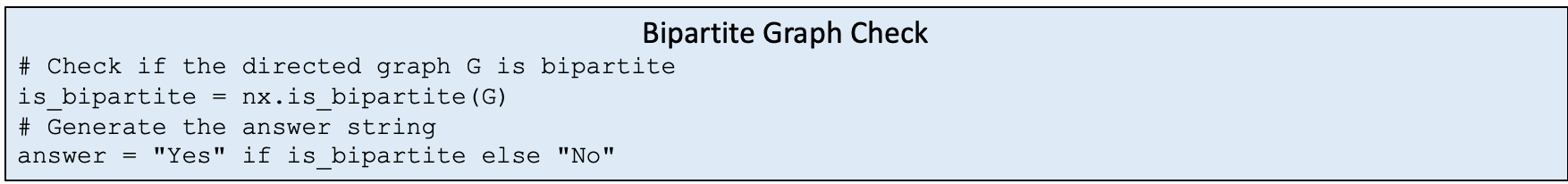}
    \caption{An example of the final code $C'$ generated for the bipartite graph check task.}
    \label{fig:bipart_exp}
\end{figure*}

\begin{figure*}
    \centering
    \includegraphics[width=\linewidth]{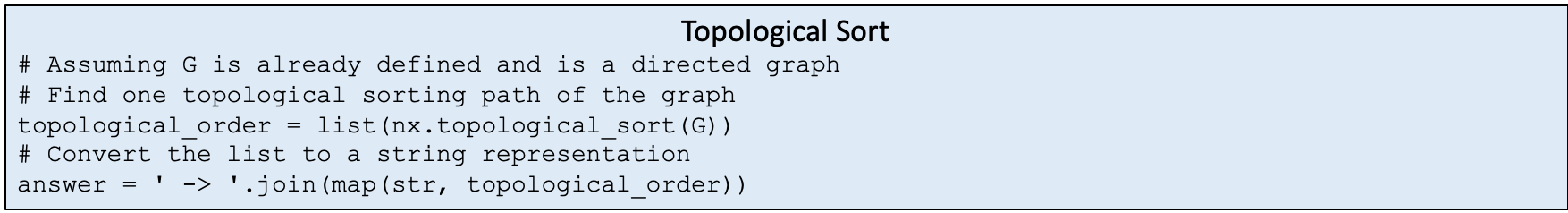}
    \caption{An example of the final code $C'$ generated for the topological sort task.}
    \label{fig:topo_exp}
\end{figure*}

\begin{figure*}
    \centering
    \includegraphics[width=\linewidth]{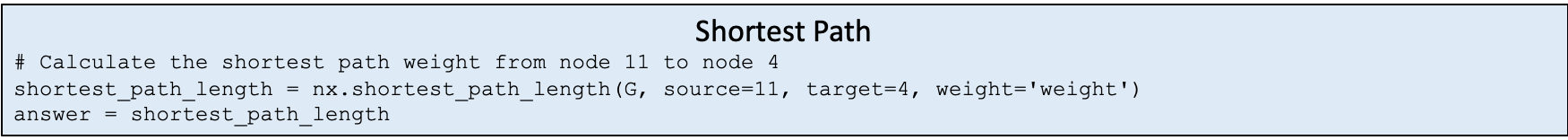}
    \caption{An example of the final code $C'$ generated for the shortest path task.}
    \label{fig:shortest_exp}
\end{figure*}

\begin{figure*}
    \centering
    \includegraphics[width=\linewidth]{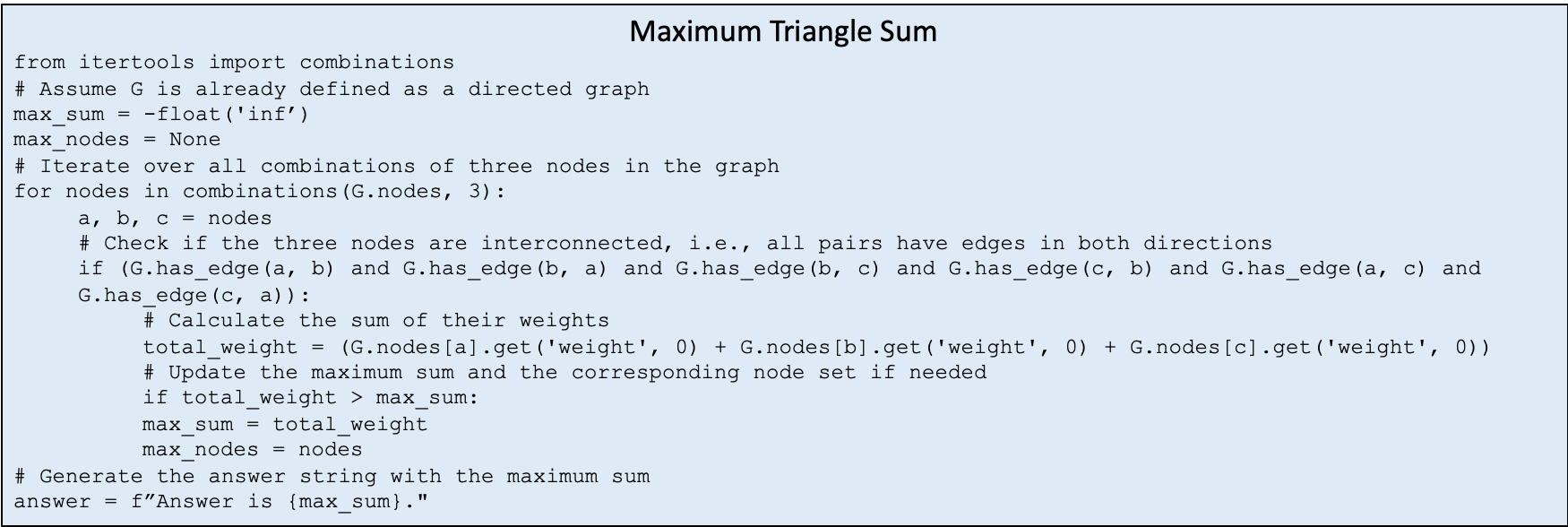}
    \caption{An example of the final code $C'$ generated for the maximum triangle sum task.}
    \label{fig:maxtri_exp}
\end{figure*}

\begin{figure*}
    \centering
    \includegraphics[width=\linewidth]{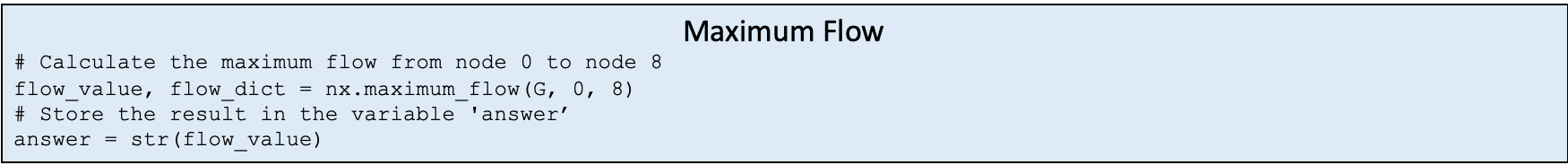}
    \caption{An example of the final code $C'$ generated for the maximum flow task.}
    \label{fig:maxflow_exp}
\end{figure*}

\begin{figure*}
    \centering
    \includegraphics[width=\linewidth]{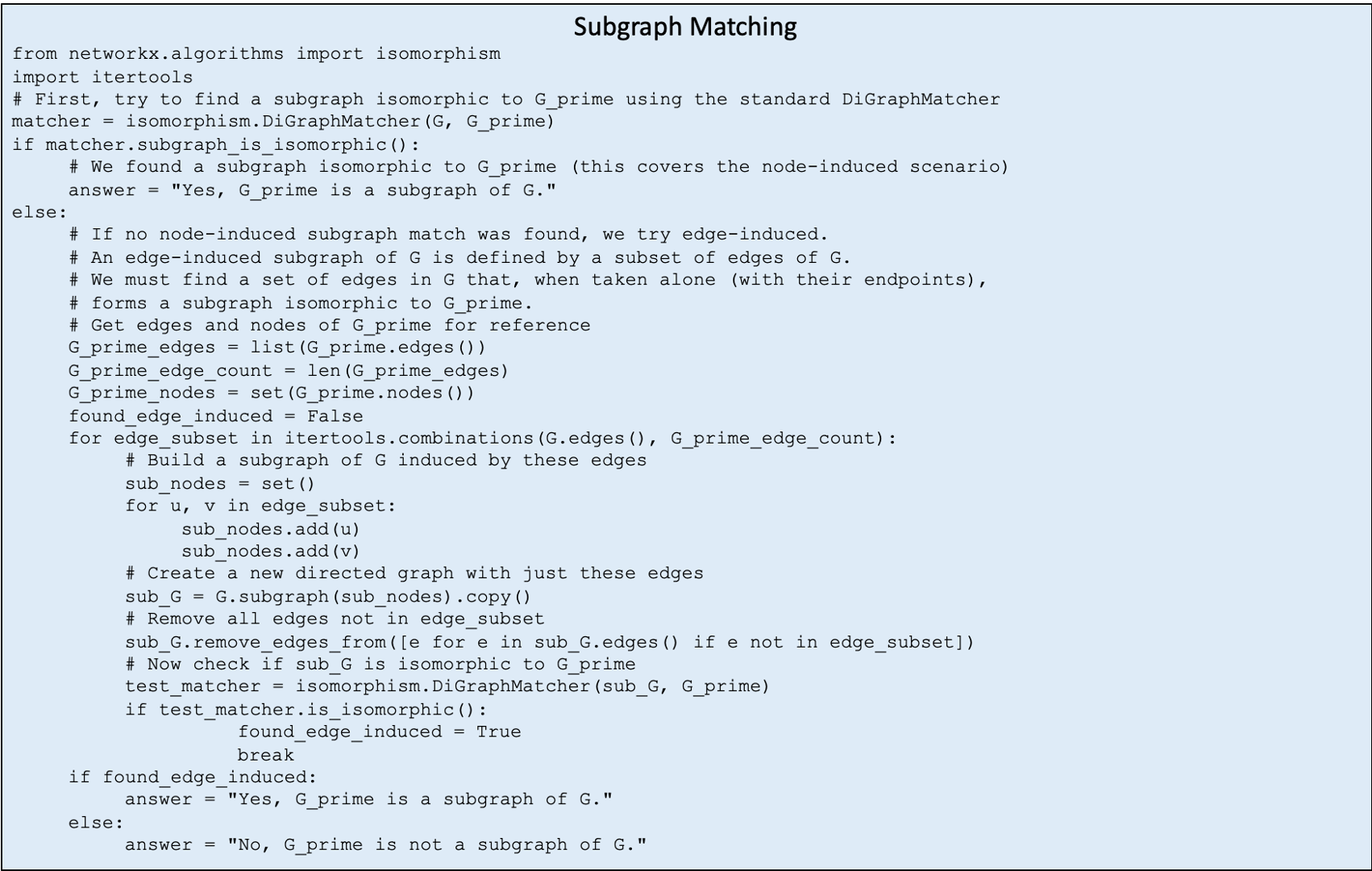}
    \caption{An example of the final code $C'$ generated for the subgraph matching task.}
    \label{fig:subgraph_exp}
\end{figure*}

\begin{figure*}
    \centering
    \includegraphics[width=\linewidth]{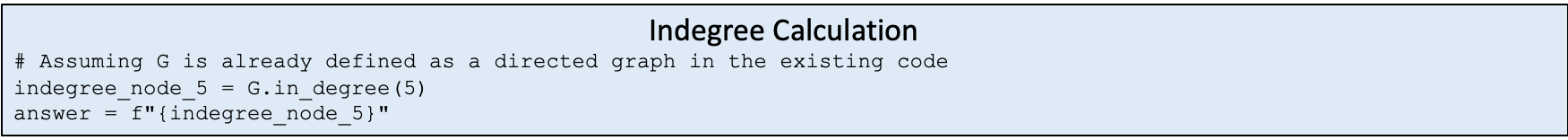}
    \caption{An example of the final code $C'$ generated for the indegree calculation task.}
    \label{fig:indegree_exp}
\end{figure*}

\begin{figure*}
    \centering
    \includegraphics[width=\linewidth]{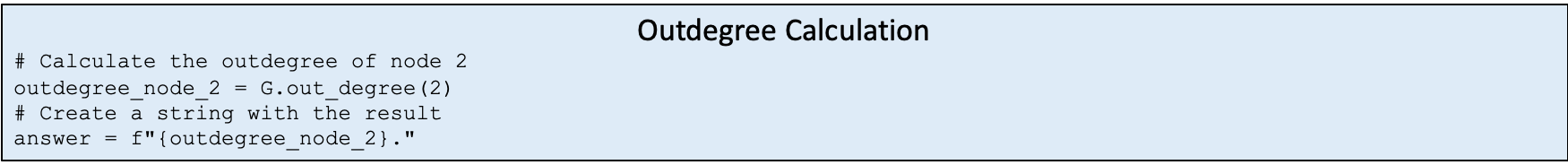}
    \caption{An example of the final code $C'$ generated for the outdegree calculation task.}
    \label{fig:outdegree_exp}
\end{figure*}

This section presents example code generated by GRRAF$_N$ for each graph reasoning task in our experiments: cycle detection (Figure \ref{fig:cycle_exp}), connectivity (Figure \ref{fig:connect_exp}), bipartite graph check (Figure \ref{fig:bipart_exp}), topological sort (Figure \ref{fig:topo_exp}), shortest path (Figure \ref{fig:shortest_exp}), maximum triangle sum (Figure \ref{fig:maxtri_exp}), maximum flow (Figure \ref{fig:maxflow_exp}), subgraph matching (Figure \ref{fig:subgraph_exp}), indegree calculation (Figure \ref{fig:indegree_exp}), and outdegree calculation (Figure \ref{fig:outdegree_exp}). \han{All these examples produce correct answers.}

\begin{figure*}
    \centering
    \includegraphics[width=\linewidth]{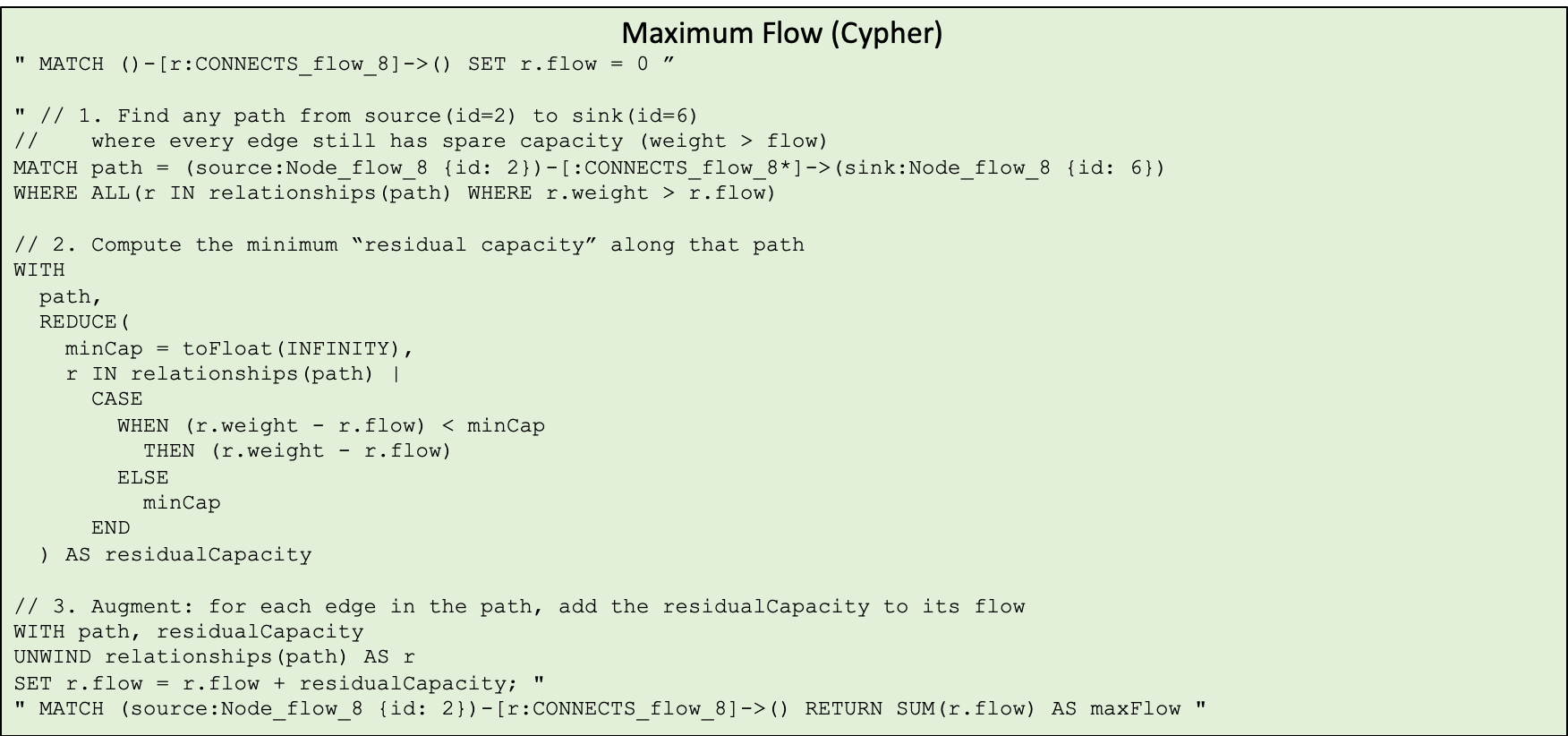}
    \caption{\han{An example of the final code $C'$ in Cypher query by GARRF$_C$ generated for the maximum flow task.}}
    \label{fig:maximum_cypher}
\end{figure*}

\begin{figure*}
    \centering
    \includegraphics[width=\linewidth]{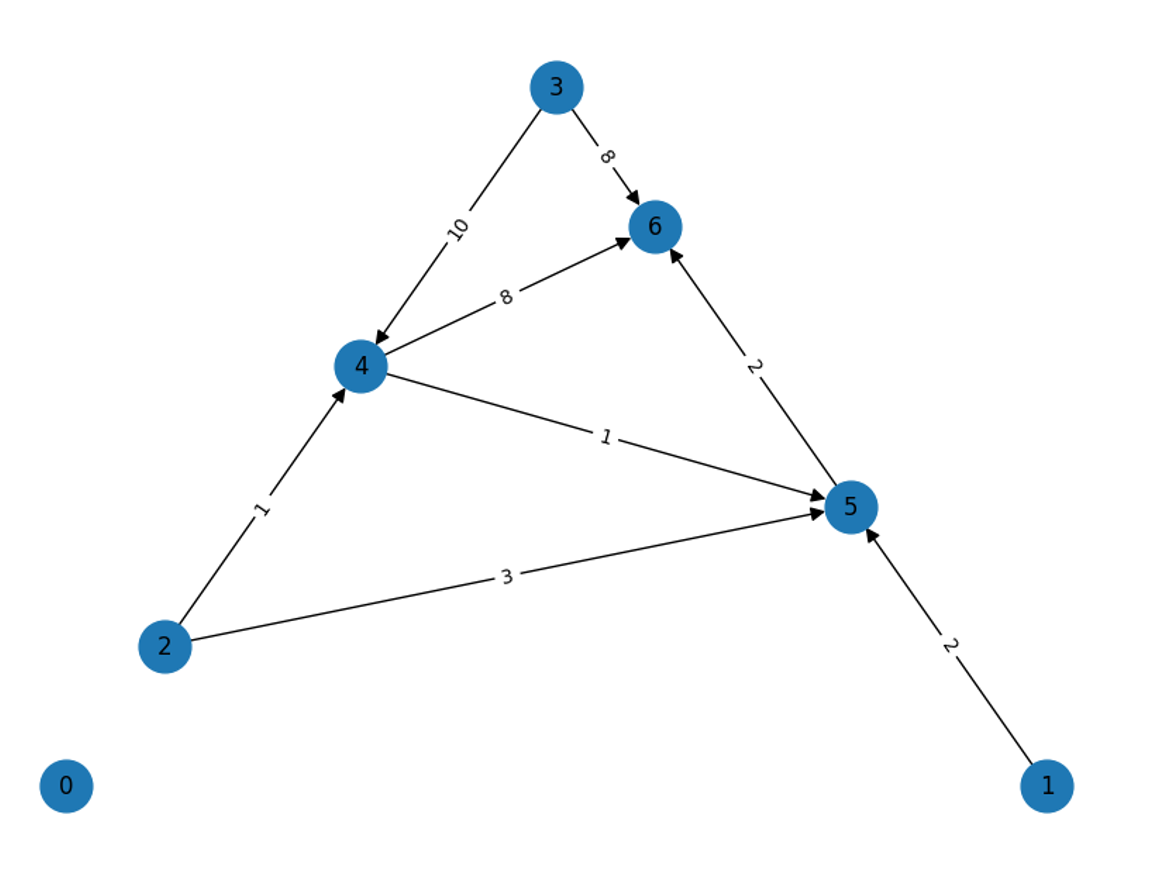}
    \caption{\han{An example directed graph with edge \hanq{weights}. The correct maximum flow from node 2 to 6 is 3 but the Cypher query in Figure} \ref{fig:maximum_cypher} returns 4 as the answer.}
    \label{fig:example_graph}
\end{figure*}

\begin{figure*}
    \centering
    \includegraphics[width=\linewidth]{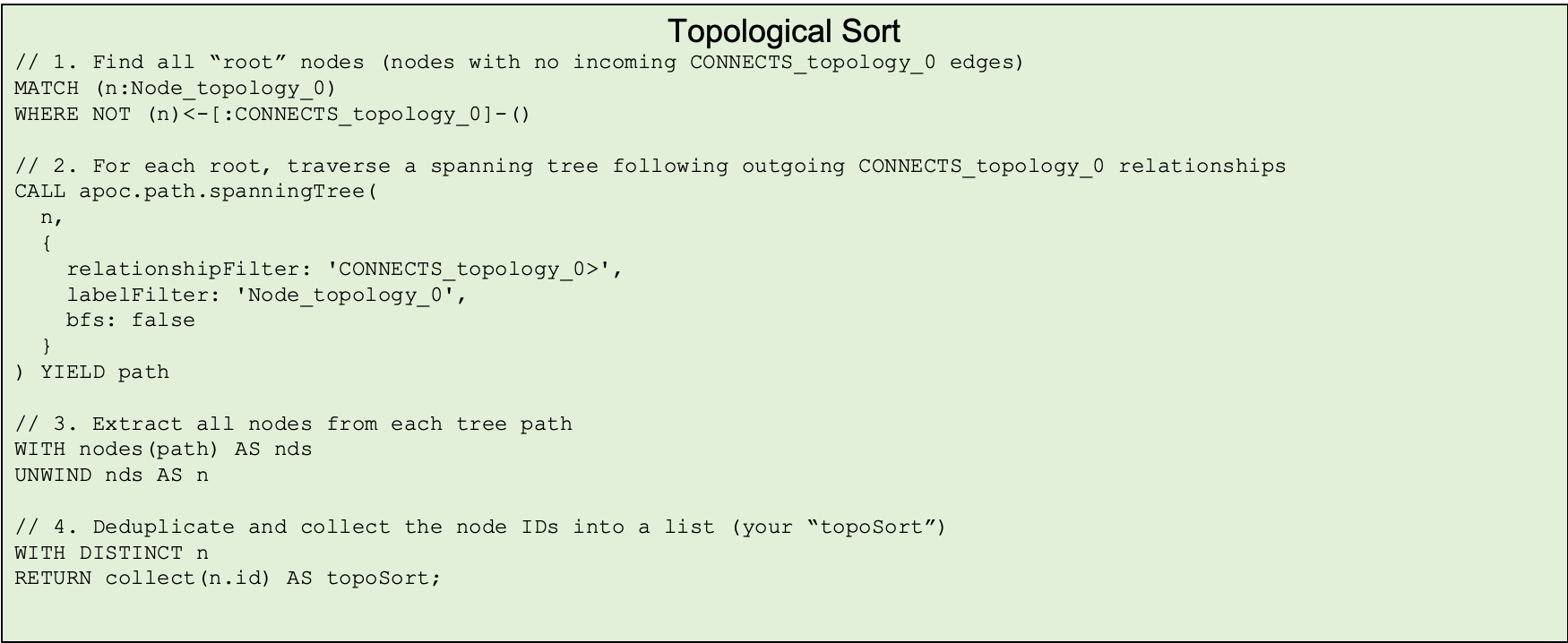}
    \caption{\han{An example of the final code $C'$ in Cypher query by GARRF$_C$ generated for the topological sort task.}}
    \label{fig:topo_cypher}
\end{figure*}

We also include in Figure \ref{fig:maximum_cypher} an example Cypher query generated by GRRAF$_C$ for the \hanq{maximum‐flow} task. Although this query attempts to implement the Ford–Fulkerson algorithm, it omits the backward residual edges, preventing any rerouting of earlier flows. Consequently, on certain edge cases (e.g., the graph in Figure \ref{fig:example_graph}), it produces incorrect results. Similarly, Figure \ref{fig:topo_cypher} shows an instance where GRRAF$_C$ generates an incorrect Cypher query for topological sorting. That query builds a spanning tree rooted at a node of zero indegree to derive the ordering—a method that is unsound and succeeds only by chance on some graphs.

\end{document}